\documentclass[runningheads]{llncs}
\usepackage{graphicx}
\usepackage{amsmath, amssymb, amsfonts}
\usepackage{bbm}
\usepackage{dirtytalk}
\usepackage{tabularx}

\begin{document}
\title{Estimating Post-OCR Denoising Complexity on Numerical Texts}
\titlerunning{Estimating Post-OCR Denoising Complexity on Numerical Texts}
%
\author{Arthur Hemmer\inst{1,2} \and
Jérôme Brachat \inst{1} \and
Mickaël Coustaty\inst{2} \and
Jean-Marc Ogier\inst{2}}
\authorrunning{A. Hemmer et al.}
%
\institute{Shift Technology, Paris \\
\email{first.last@shift-technology.com} \and
La Rochelle Université, L3i \\
Avenue Michel Crépeau, 17042 La Rochelle, France \\
\email{first.last@univ-lr.fr}
}
\maketitle              
\begin{abstract}
Post-OCR processing has significantly improved over the past few years. However, these have been primarily beneficial for texts consisting of natural, alphabetical words, as opposed to documents of numerical nature such as invoices, payslips, medical certificates, etc. To evaluate the OCR post-processing difficulty of these datasets, we propose a method to estimate the denoising complexity of a text and evaluate it on several datasets of varying nature, and show that texts of numerical nature have a significant disadvantage. We evaluate the estimated complexity ranking with respect to the error rates of modern-day denoising approaches to show the validity of our estimator.

\end{abstract}
\section{Introduction}
Optical character recognition (OCR) is the process of converting text from the visual domain into machine-readable text. It plays an essential role in bridging the gap between the physical and the virtual. Many businesses, governments and individuals rely on OCR to effectively manage documents of various types.

While OCR accuracy has improved greatly over the years, it remains an active area of research. In industries such as finance and insurance, high OCR accuracy has become crucial as it is used in fraud detection systems. These systems often work on semi-structured, scanned documents such as invoices, medical certificates, bank statements, etc. While the accuracy of modern-day OCR might be sufficient for information retrieval use cases, it falls short for these fraud detection use cases where wrong predictions can cause many false positives. These systems often rely on one or a few fields of highly specific nature from an array of documents. Small amounts of OCR noise occurring on these fields has a multiplicative, negative impact on the end-to-end accuracy of such fraud detection systems.

\begin{figure}[tbh]
    \centering
    \includegraphics[width=250pt]{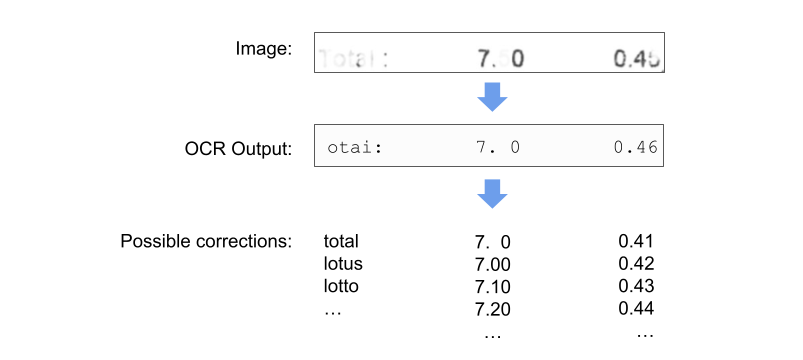}
    \caption{Visualisation of post-OCR correction process. Business information such as amounts are harder to denoise due to the possibility of all orthographic neighbors} \label{examplecorrections}
\end{figure}

In order to combat noisy OCR output, one often uses OCR post-processing methods \cite{chiron2017icdar,rigaud2019icdar}. A classical approach combines a model of the typical errors that the OCR makes with a prior about the text that is processed consisting of a vocabulary with corresponding word frequencies. While this approach works well for natural language texts where the vocabulary is finite and well-defined, it is less effective for texts coming from business documents with numerical words such as dates, amounts, quantities, invoice numbers, etc. which are not contained in typical natural language vocabularies. An example of this is shown in Figure \ref{examplecorrections}, where a noisy reading of a dictionary word has few orthographically close corrections whereas numerical words cannot rely on this technique as all orthographic neighbours are equally likely. Most OCR post-processing approaches of aforementioned kind completely ignore these numerical words \cite{dannells2020swedishocr,taghva2001ocrspell,thawani2021representing} or treat them as normal words \cite{dutta2022pnrank}, leaving them prone to errors.


\vspace{-0,3cm}

A more modern approach to OCR post-processing is to consider it as a sequence-to-sequence (\say{Seq2Seq}) problem, which is already widely researched for tasks such as translation and speech recognition. 
This type of approach has been boosted by deep-learning models and large parallel corpora. While these methods achieve state-of-the-art performance on OCR post-processing tasks \cite{chiron2017icdar,rigaud2019icdar}, much like the classical vocabulary-based approaches, these language models are ineffective on numerical words \cite{mitchell2009language,thawani2021representing}. Many methods do not make any distinction when considering numerical versus non-numerical words \cite{dutta2022pnrank,jatowt2019post,pham2021candidate}. As such, some of them report that the majority of non-word errors come from tokens containing numbers \cite{jatowt2019post}. In some cases, tokens containing punctuation and/or numbers are filtered out of the dataset entirely \cite{dannells2020swedishocr,taghva2001ocrspell}. In other cases, the presence of non-alphanumeric characters is even considered as an important positive indicator for detecting erroneous words \cite{dannells2020swedishocr}. 


While the overall denoising performance has increased with these seq2seq approaches, we hypothesize that this improvement has been biased towards natural language, leaving datasets of more numerical nature untouched. The aim of this article is to quantify and compare the post-OCR denoising complexity of various datasets of both numerical and non-numerical nature. We do this by simulating textual noise and estimating the complexity by computing the performance of a simple denoising method under optimal conditions. Furthermore, we establish the real-world applicability of these estimates by comparing it to the performance of two cutting-edge post-OCR processing approaches under more realistic noise conditions. With these insights, we hope to shed more light on the strengths and weaknesses of modern-day post-OCR processing approaches and provide directions for future research. To summarize, in this paper we propose the following elements:
\begin{enumerate}
    \item A formalization for estimating the OCR denoising complexity of a dataset
    \item An evaluation of these estimates with respect to the performance from post-OCR processing approaches in more realistic settings
\end{enumerate}

\vspace{-0,3cm}
\section{Related Work} \label{relatedwork}

Early work on estimating the denoising complexity of texts subjected to noise looked at the impact of the size of the vocabulary on the number of real-word errors \cite{peterson1986note}, which are erroneous words that also occur in the vocabulary. They show that the fraction of these errors increases rapidly up to $13\%$ for $100,000$ words and then increases much more slowly to $15\%$ for $350,000$ words. While the conclusion states that smaller word lists are beneficial, this is only true for the real-word error rate. Hence, this conclusion was rightfully challenged \cite{damerau1989examination} by showing that decreasing the size of the dictionary also increases the number of non-word errors, which are wrongly corrected words because the correct word was not in the vocabulary. The example they give is if \textit{coping} were omitted from the vocabulary, the 4 misspellings of \textit{copying} would be detected. However, the 22 correct uses of \textit{coping} would be flagged as misspelled.

While these experiments look specifically at the impact of the size of the vocabulary, two important elements are not taken into consideration: the syntactic distribution of the words inside a vocabulary and the underlying noise model. For example, a small vocabulary consisting of words that are all within one edit distance from each other (numerical words) will be much harder to denoise than a large vocabulary where all words are within multiple character edits (natural words). As for the noise model, both \cite{damerau1989examination,peterson1986note} assume a uniform probability for each edit operation (transpose, add, remove, substitute) whereas there are factors that skew this distribution such as keyboard layout and phonological ambiguities. It should also be noted that this previous work was conducted in the context of human typing errors whereas it has been shown that human typing errors and OCR errors do not have the same characteristics \cite{jatowt2019deep}. Spelling errors typically generated by humans do not correspond to the noise that an OCR would introduce. For example, $63\%$ of human misspellings occur in short words (of length 4 or less) whereas this is only $42.1\%$ for OCR errors. To our knowledge, there is no prior work on estimating the denoising complexity of post-OCR processing approaches.

\section{Background}
The correction of spelling errors, whether they originate from humans or OCR software, has been a widely researched topic. Formally speaking, the goal is to find the original sequence $w$ from a noisy observed sequence $o$, given a probabilistic model $p(w|o)$. As such, we denote the estimator of $w$ as $\hat{w}$ such that:

\begin{equation}\label{optestim}
    \hat{w}(o) := \arg\max_{w} p(w|o)
\end{equation}

While $w$ and $o$ can be any type of sequence (characters, words, sentences, paragraphs, etc.), it is typically considered at the character or word level due to limits in computational complexity. The parameters of this model have historically been estimated either by decomposing according to the noisy channel model or directly using more advanced sequence-to-sequence approaches, both of which are discussed separately below.

\subsection{Noisy Channel Model}
First works on error correction \cite{church1991probability,kemighan1990spelling} estimated the parameters for $p(w|o)$ by applying the noisy channel model \cite{shannon1948mathematical}. This works by applying Bayesian inversion to Equation \ref{optestim} and dropping the denominator as it does not impact the result of the $\arg\max$ function. As this approach works on a word level, the $\arg\max$ is taken with respect to a finite vocabulary $\mathcal{V}$ where $w \in \mathcal{V}$.


\begin{equation}
    \hat{w}(o) = \arg\max_{w \in \mathcal{V}} p(o|w)p(w)
    \label{postprocessing}
\end{equation}

In this form, $p(o|w)$ and $p(w)$ are often referred to as the noise model and the language model (or prior), respectively. The noise model denotes the probability of observing a noisy sequence $o$ from $w$. More often than not, there is not enough data available to directly compute $p(o|w)$. Instead, the noise is decomposed in individual character edits such as substitutions, insertions and deletions.

In the simplest case, the prior $p(w)$ consists of individual word probabilities. These can be estimated directly from the training data or come from auxiliary corpora. However, a single-word prior is restricted in the amount of information it can provide. 
To solve this issue, many approaches also take into account the surrounding context of a word where the language model becomes a word $n$-gram model or a more capable neural network-based language model. Using $\hat{w}_i$ to denote the $i$-th denoised word in a sequence gives us:

\begin{equation}
    \hat{w}_i(o) = \arg\max_{w \in \mathcal{V}} p(o|w)p(w|\hat{w}_{i-1}, \hat{w}_{i-2}, \ldots, \hat{w}_{i-n})
    \label{ngramlm}
\end{equation}

While this enables a noisy channel denoiser to take into account the context of a word, it also introduces a new potential source of errors as the prior is conditioned on the previous estimates for $\hat{w}$ and not the true words $w$. An erroneous prediction for $\hat{w}_i$ can have a negative impact on the prior. \textit{Beam search} \cite{dahlmeier2012beam} is often used to counter this problem. Instead of relying on a single prediction for each word, beam search keeps track of a top (fixed) number of candidates at each prediction step and computes the $\arg\max$ for each of these candidates at the next prediction, and so on.

\subsection{Sequence-to-Sequence Models}
One can also use more capable methods to directly estimate $p(w|o)$ instead of decomposing it into a noise and language model. This approach is widely used in machine translation, where it is referred to as neural machine translation (NMT) when using deep-learning methods, and has also been shown to work well on text error correction \cite{nguyen2020neural,ramirez2022post,soper2021bart}. It works by using an encoder-decoder \cite{cho2014encdec} architecture, where the encoder takes the whole noisy input sequence and encodes it into a fixed length vector. A decoder is then conditioned on this vector and its own previous outputs to generate subsequent words in an autoregressive manner. This gives us the following estimator:

\begin{equation}
    \hat{w}_i(o) = \arg\max_{w \in \mathcal{V}}p(w|o,\hat{w}_{i-1}, \hat{w}_{i-2}, \ldots, \hat{w}_{i-n})
    \label{encdeclm}
\end{equation}



Similar to the n-gram approach, the autoregressive nature of these models is a potential source for errors. In the same manner, beam search can also be used for these direct estimators to overcome such errors.

\section{Denoising Complexity} \label{dencompl}
As discussed, there are various approaches to post-OCR processing. However, our hypothesis is that the frequency of numerals has a significant impact on the denoising complexity of a text, regardless of the used denoising approach. As such, we devise a simple method for quantifying the complexity of a text. We consider the noisy channel decoder from Equation \ref{postprocessing} under optimal conditions meaning that the denoiser has access to the true noise model and prior for a given text. For the prior, we use a unigram word frequency prior $p(w)$. The following subsection provides more details on the noise model before getting to the estimation of the complexity at last.

\subsection{Noise Model}
The noise model, $p(o|w)$, is a substitution-only noise model that we denote with $\pi$. While substitution-only is a simplification of reality where OCR errors can also contain insertion and deletion errors, it has been shown that the majority of errors consist of character substitutions \cite{jatowt2019deep}. Furthermore, we challenge this simplification in subsection \ref{sectapplicability}, where we also include insertions and deletions in more realistic evaluation scenarios.

We compute the probability of obtaining word $w$ from an observed word $o$ under noise model $\pi$ by taking the product of the individual character confusion probabilities. Here, $w^i$ and $o^i$ denote the character of token $w$ and $o$ at index $i \in \{1, 2, ... , n\}$ where $n$ is the length of the token.

\begin{equation}
    p_\pi(o|w) = \prod_{i=1}^{|w|} p_{\pi}(o^i|w^i) 
    \label{noisemodel}
\end{equation}

Where $|w|$ denotes the length of word $w$, and $p_{\pi}(o^i|w^i)$ the probability of observing a character $o^i$ given a character $w^i$ under noise model $\pi$. Since we are considering only substitutions, we will only consider $o$'s that have the same length as $w$ and vice versa. In other words, if $|w| \neq |o|$, then $p(o|w) = 0$.

Throughout the experiments we consider two noise models: a uniform noise model $\pi_{\epsilon}$ and a more realistic OCR noise model $\pi_{ocr}$. Given an alphabet $\mathcal{A}$ of possible characters, the uniform noise model $\pi_{\epsilon}$ has probability $\epsilon$ of confusing a character and probability $1-\epsilon$ of keeping the same character. Within the substitution probability, each character has probability of $\epsilon/(|\mathcal{A}| - 1)$ for being substituted.

The second noise model, $\pi_{ocr}$, is estimated from the English part of the ICDAR 2019 OCR post-processing competition dataset \cite{rigaud2019icdar}. The dataset contains a total of 243,107 characters from over $200$ files from IMPACT\footnote{https://www.digitisation.eu/}. The purpose of this noise model is to evaluate our estimator in a more realistic setting, as in practice OCR programs tend to have sparse confusion probabilities. For example, this means that when a mistake is made on a character such as \say{1}, it is most often confused for visually similar characters such as \say{i}, \say{t} and \say{l} and not so often by \say{8} or \say{Q}.

\subsection{Complexity Estimator}
Finally, using the previously described noise model and noisy channel model, let us denote the denoising complexity of a dataset under noise model $\pi$ as $\Theta_\pi$. We define $\Theta_\pi$ by considering the accuracy of the optimal denoising algorithm under the noisy channel model with a unigram prior. The denoising complexity is estimated by taking the expectation of the number of errors according to the noise model.

\begin{equation}
    \Theta_{\pi} = \mathbb{E}_{o,w \sim \pi}[ \mathbbm{1}\{ w \neq \hat{w}_\pi(o) \}]
    \label{theta}
\end{equation}

Where we use $\hat{w}_\pi$ according to Equation \ref{optestim}. We estimate it by sampling words $w \sim p(w)$ from our dataset and obtaining $o$ by applying the noise model such that $o \sim \pi(w)$. An important advantage of our estimator is that it is computationally simple and highly parallelizable.


The intuitive interpretation of $\Theta_\pi$ is that it is the expected probability of picking an incorrect word given its noisy observation. It is the word error rate of a unigram denoising approach, but under optimal conditions. Having the true prior allows us to compare complexities between different datasets, as we rule out any variance that comes from having a sub-optimal estimate of the prior. In other words, having the optimal prior for a given dataset allows us to estimate and compare exactly our quantity of interest.

\section{Experiments}
Using our complexity estimator, we devise a ranking of denoisability of textual datasets of varying nature. Following this, we evaluate this ranking with respect to the performance of more advanced denoisers in a more realistic noise setting. 

In all experiments, we evaluate a total of five datasets. We chose two datasets of more numerical nature FUNSD \cite{jaume2019funsd} and SROIE \cite{huang2019sroie}, and three datasets of more alphabetical nature OneStopEnglish \cite{vajjala2018onestopenglish}, KleisterNDA  \cite{gralinski2020kleister} and IAM \cite{marti2002iam}. Each dataset is tokenized using the SpaCy\footnote{\textit{en\_core\_web\_sm} from SpaCy v3.4.4 from https://spacy.io/} tokenizer. 

\begin{table}[tbh]
\centering
\caption{The datasets used in the experiments along with relevant statistics}
\label{tabdatasets}
\begin{tabularx}{0.9\textwidth}{lrrrrrrl}
\hline
Dataset        & Documents & \textbar{}$\mathcal{V}$\textbar{} & \textbar{}$\mathcal{V}_\#$\textbar{} & \textbar{}$\mathcal{V}_\alpha$\textbar{} & $p(\mathcal{V}_\#)$ &  $p(\mathcal{V}_\alpha)$ & Document Type      \\ 
\hline
FUNSD          & 149          & 5503                              & 1477                                 & 3634  & 0.138  & 0.617 & Forms              \\
IAM            & 1277         & 11598                             & 339                                  & 9776  & 0.007  & 0.841 & Handwritten lines  \\
KleisterNDA    & 254           & 12418                              & 1988                                  & 9850  & 0.015  & 0.835 & Legal documents    \\
OneStopEnglish & 453          & 15807                            & 710                                  & 14791 & 0.016  & 0.847 & Educational texts              \\
SROIE          & 626          & 11397                             & 7176                                 & 3838  & 0.246  & 0.480 & Receipts           \\
\hline
\end{tabularx}
\end{table}

We use $\mathcal{V}$ to denote the vocabulary which represents the set of words present in a dataset. In addition, we use $\mathcal{V}_\#$ to denote the numerical vocabulary which is the subset of words containing at least one number, and $\mathcal{V}_\alpha$ to denote the alphabetical vocabulary which is the subset of words containing only letters. Note that $\mathcal{V}_\# \cap \mathcal{V}_\alpha = \emptyset$, but $\mathcal{V}_\# \cup \mathcal{V}_\alpha$ is not necessarily equal to $\mathcal{V}$ since we do not count punctuation and special characters in the alphabetical vocabulary.
All datasets along with some descriptive statistics can be found in Table \ref{tabdatasets}. We also included $p(\mathcal{V}_\#)$ and $p(\mathcal{V}_\alpha)$ which are the frequencies of the words in that vocabulary with respect to the whole dataset. As can be seen, FUNSD and SROIE have significantly higher frequencies of numerical words than IAM, Kleister-NDA and OneStopEnglish.

\subsection{Denoising Complexity} \label{sectdenoiscomplex}
Using previously described noise models and datasets, we estimate the complexity by sampling $10^6$ words according to $p(w)$ and apply random substitutions according to the noise model to obtain observed word $o$. We then use the sampled ($w$,$o$) pairs to estimate the complexity according to equation \ref{theta}. To estimate the complexity at varying degrees of noise, we gradually interpolate the noise from the character confusion matrix $M_\pi$ with the identity matrix $I$ using a parameter $\gamma \in [0,1]$ such that $M_{noise} = \gamma M_{\pi} + (1 - \gamma)I$.

\begin{figure}[tbh]
    \centering
    \includegraphics[width=420pt]{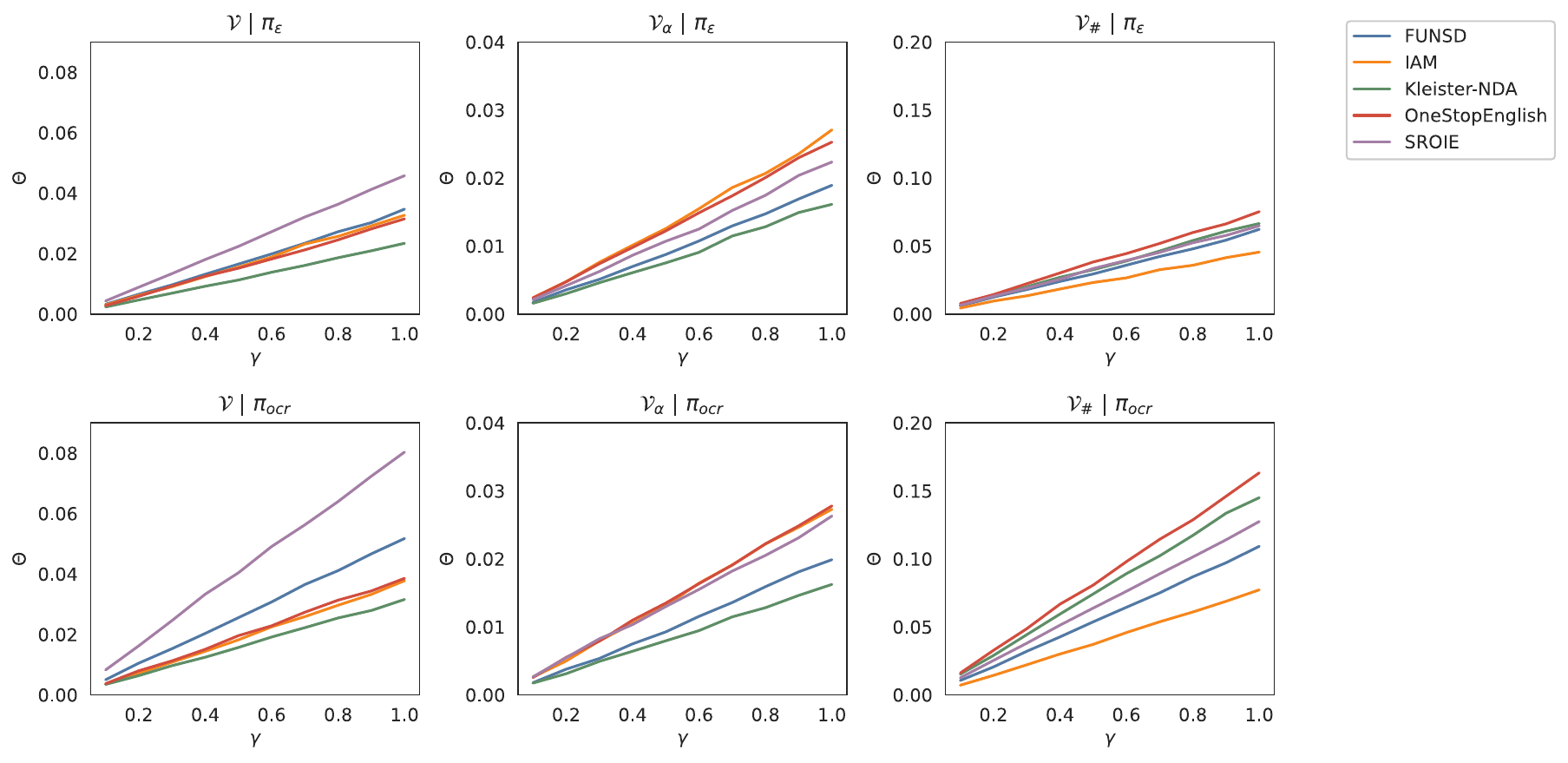}
    \caption{Denoising complexity $\Theta$ for increased noise levels $\gamma$ under noise models $\pi_{\epsilon}$ (left) and $\pi_{ocr}$ (right).} \label{basecomplexity}
\end{figure}

In our experiments we set $\epsilon=0.07$ for the uniform noise $\pi_\epsilon$. We chose this value because it aligns with the average confusion probability of the estimated OCR noise model. All results are computed for $\gamma$ increments of $0.1$ starting from $0.1$ up to $1.0$.  We found that these increments gave us a good balance between computing time and visualisation value. For each noise model we also estimate the complexity on alphabetical words ($\mathcal{V}_\alpha$) and numerical words ($\mathcal{V}_\#$) specifically. The results can be found in Figure \ref{basecomplexity}.

Our primary observation is that the complexity ranking for $\mathcal{V}$ is preserved between the two noise models $\pi_{\epsilon}$ and $\pi_{ocr}$, and increases linearly with respect to $\gamma$. Under both noise models, the two datasets with the largest frequencies of numerical words (SROIE and FUNSD) have the highest complexity. Going from $\pi_\epsilon$ to $\pi_{ocr}$, their complexity increases, going from $0.047$ to $0.084$ for SROIE and from $0.042$ to $0.071$ for FUNSD, respectively. The steep increase in complexity for the numerical datasets can be mostly attributed to the higher average confusion probability for numbers for the OCR noise ($0.14$) compared to the uniform noise ($0.07$), combined with the significantly higher frequency of numerical words compared to the other datasets (see column $p(\mathcal{V}_\#)$ in Table \ref{tabdatasets}). The other three, mostly alphabetical datasets show overall lower values for $\Theta$, implying a lower denoising complexity. IAM and OneStopEnglish have close estimates under both the uniform and OCR noise models, though slightly higher for IAM in both cases.

Looking at the complexity estimates of the numerical vocabulary $\mathcal{V}_\#$, we observe them to be much higher than for the other vocabularies, even under the uniform noise model. Interestingly, the complexity ranking changes between the different vocabularies. Considering the complexity ranking of the numerical vocabulary, both OneStopEnglish and Kleister-NDA have similar or higher estimates than FUNSD and SROIE. A qualitative analysis of the results shows this to be due to the nature of numerical words in alphabetical datasets. In these datasets, numerical words used in natural language are often single numbers used for single counts (such as \say{Bob gave me 2 euros}), or numbers with low variation such as year numbers (\say{2007}, \say{2008}) and large rounded numbers (\say{10,000}, \say{20,000}, etc). This is in contrast with FUNSD and SROIE, where numerical words consist primarily of amounts or dates which are longer, more diverse number sequences and thus slightly easier to denoise given that the numerical vocabulary does not cover all possible amounts. Note that while \mbox{OneStopEnglish} shows a high complexity for the numerical vocabulary, its overall complexity remains lower than FUNSD and SROIE, due to the lower frequency of numerical words in the dataset.

\subsection{Applicability}\label{sectapplicability}
The results from subsection \ref{sectdenoiscomplex} show us a relative denoising complexity for various datasets. However, when defining our estimator, we made several simplifying assumptions in order to compute this complexity. In this second part of our experiments, we wish to evaluate the applicability of our complexity estimate using a more realistic noise model as well as more advanced denoising methods.

First, we extend the noise model to also include insertions and deletions. The insertion and deletion probabilities for the OCR noise are estimated from the ICDAR 2019 OCR post-processing competition dataset\cite{rigaud2019icdar}. To both the uniform and OCR noise models, we add the possibility for insertion and deletion with probabilities of $0.03$ and $0.04$ respectively. 

Second, we evaluate the performance of 3 state-of-the-art denoising methods of the encoder-decoder architecture type. It consists of 2 transformer approaches ByT5 \cite{xue2022byt5} and BART \cite{soper2021bart}, and one Recurrent Neural Network (RNN) trained using OpenNMT \cite{klein-etal-2017-opennmt}. ByT5 is a version of T5 \cite{raffel2020t5} where the tokens are characters (bytes) instead of the usual SentencePiece tokens, which makes it more suitable for text denoising. Both transformer models were initialized from their publicly available pretrained weights (base) and were fine-tuned using an Adam optimizer with a learning rate of $0.0001$ for $10$ epochs. The RNN is trained on character sequences where the characters are separated by spaces and the words separated by \say{@}. For coherence, we used the same hyperparameters as \cite{nguyen2020neural} for denoising OCR errors. 

The data is preprocessed by concatenating all the datasets and splitting documents on spaces, of which the resulting token sequences are then used to create target sequences of at most 128 characters in length. The noisy sequences are generated by applying the noise model on the target sequences. To handle longer documents during evaluation, we split the input text again on spaces and denoise sequences of at most 128 characters at a time, after which they are concatenated again to form the final denoised prediction for a document. While it is technically possible for a word such as \say{article} to be noised into two separate words \say{art icle} and then split between evaluation sequences, we consider this to be rare enough as to not impact the results and at worst impact all denoisers equally. A separate model was trained for the uniform and OCR noise.

Finally, we compute the performance of each denoising method by computing the word error rate (WER) between the predicted output and the ground truth. In this case we use the non-normalized error rate which is the edit distance between the two tokenized sequences divided by the number of tokens in the ground truth sequence. We also include a baseline which is the WER that is computed from the original and unprocessed noisy sequence. This is to evaluate the relevance of our estimator.

The results are shown in Table \ref{full_noise_wer}. Our initial observation is that the baseline WER ranking does not follow the ranking of our complexity estimation, nor does it correspond to the WER of the other denoisers. Under uniform noise, SROIE and FUNSD show fewer errors than Kleister-NDA for the baseline. Under OCR noise, IAM has the lowest error rate, after Kleister-NDA and OneStopEnglish which have similar error rates.

We observe the error rates for the numerical datasets (FUNSD and SROIE) to be higher than the others for both BART and ByT5. While this is not the case for the OpenNMT denoiser, it should be noted that it has poor overall performance as it performs similar or worse than the baseline, with the exception being the Kleister-NDA dataset. As the Kleister-NDA dataset was significantly larger than the others, we suspect that the OpenNMT denoiser overfit on this dataset. Although BART achieves the lowest WER on Kleister-NDA, ByT5 has on average the lowest WER. Most notably, the gap in WER between the two numerical datasets FUNSD and SROIE is smaller for ByT5 under both noise models (uniform: +0.01, OCR: -0.01), whereas BART has consistently more difficulty with FUNSD (uniform: +0.05, OCR: +0.06). On the non-numerical datasets, Kleister-NDA has consistently the lowest WER, with IAM and OneStopEnglish having nearly the same WER for BART and ByT5 under both uniform and OCR noise.

Compared to our complexity estimates, we do note some inconsistencies. First, FUNSD and SROIE are much closer in terms of their WER than their complexity estimates. For BART, FUNSD even has 5 and 6 percent points higher WER than SROIE under uniform and OCR noise respectively. While still close and much higher than the non-numerical datasets, we suspect this difference to come from the amount of training data which is three times higher for SROIE (96k tokens) compared to FUNSD (26k tokens). In addition, the receipts from SROIE are very homogenous and contain many longer recurring subsequences such as \say{gardenia bakeries (kl) sdn bhd (139386 x) lot 3} and \say{payment mode amount cash}. Furthermore, this advantage for SROIE seems to be unique to BART, as the WER under OCR noise for ByT5 shows a higher value for SROIE than for FUNSD. We suspect that the sub-word token approach used for BART is better able to model these longer recurring sequences from SROIE compared to the character-based approaches.


\begin{table}[tbh]
\centering
\caption{WER for the denoisers under the full noise model including insertions and deletions. ONMT = OpenNMT. \textbf{Bold} indicates the lowest WER for a given denoiser.}
\label{full_noise_wer}
\begin{tabularx}{\textwidth} {
| m{7em} |
>{\centering\arraybackslash}X 
>{\centering\arraybackslash}X 
>{\centering\arraybackslash}X 
>{\centering\arraybackslash}X |
>{\centering\arraybackslash}X 
>{\centering\arraybackslash}X
>{\centering\arraybackslash}X
>{\centering\arraybackslash}X |
}
\hline
 & \multicolumn{4}{c|}{Uniform} & \multicolumn{4}{c|}{OCR} \\
Dataset & Baseline & BART & ByT5 & ONMT & Baseline & BART & ByT5 & ONMT \\
\hline
FUNSD          &     0.57        & 0.30        & 0.26        &    0.61        &     0.55 & 0.36 & 0.28 &    0.60 \\
IAM            &     0.54        & 0.22        & 0.20        &    0.56        &     \bf{0.48} & 0.26 & 0.21 &    0.45 \\
Kleister-NDA   &     0.61        & \bf{0.08} & \bf{0.11} &    \bf{0.23} &     0.55 & \bf{0.10} & \bf{0.10} &    \bf{0.28} \\
OneStopEnglish &     0.60        & 0.21        & 0.19        &    0.68        &     0.54 & 0.25 & 0.21 &    0.53 \\
SROIE          &     \bf{0.51} & 0.25        & 0.25        &    0.55        &     0.57 & 0.30 & 0.29 &    0.52 \\
\hline
\end{tabularx}
\end{table}

\section{Conclusion}
We introduced a post-OCR error denoising complexity estimator, and evaluated its validity by comparing it to more complicated approaches in a more realistic setting. Furthermore, we also evaluated the complexity of specifically alphabetical and numerical words, to highlight the contribution of words of varying nature to the to the overall denoising complexity when they are sufficiently frequent. Future extensions of this work could look at the impact of using OCR word/character confidence distributions, which are sometimes available and exploited by denoising algorithms. Additionally, it would be interesting to research denoising approaches that specifically improve the denoising complexity of numerical datasets, as this would be most useful in industries relying on documents of primarily numerical nature.

%
%
\bibliographystyle{splncs04}
\bibliography{bibliography}

\end{document}